\documentclass{article}
\usepackage[final]{nips_2017}
\usepackage[utf8]{inputenc} 
\usepackage[T1]{fontenc}    
\usepackage{hyperref}       
\usepackage{url}            
\usepackage{booktabs}       
\usepackage{amsfonts}       
\usepackage{nicefrac}       
\usepackage{microtype}      
\usepackage{graphicx}
\usepackage{amsmath}
\usepackage{subfig}
\usepackage{xcolor}

\usepackage{float}
\usepackage{listings}
\title{NoveltyRank: \\
A Retrieval-Augmented Framework for Conceptual Novelty Estimation in AI Research
}

\author{
  Zhengxu Yan$^{1,*}$ \quad
  Han Li$^{2,*}$ \quad
  Yuming Feng$^{2,*}$ \\
  $^{1}$Department of Computer Science, Stanford University \\
  $^{2}$Department of Electrical Engineering, Stanford University \\
  \texttt{jasonyan@stanford.edu} \quad
  \texttt{cli27@stanford.edu} \quad
  \texttt{yumingf@stanford.edu} \\
  {\small *Equal contribution.}
}

\makeatletter
\renewcommand{\@noticestring}{}
\makeatother

\begin{document}


\maketitle
\begin{abstract}
\vspace{-1em}
The accelerating pace of scientific publication makes it difficult to identify truly original research among incremental work. We propose a framework for estimating the \textit{conceptual novelty} of research papers by combining semantic representation learning with retrieval-based comparison against prior literature. We model novelty as both a binary classification task (novel vs.~non-novel) and a pairwise ranking task (comparative novelty), enabling absolute and relative assessments. Experiments benchmark three model scales, ranging from compact domain-specific encoders to a zero-shot frontier model. Results show that fine-tuned lightweight models outperform larger zero-shot models despite their smaller parameter count, indicating that task-specific supervision matters more than scale for conceptual novelty estimation. We further deploy the best-performing model as an online system for public interaction and real-time novelty scoring.
\end{abstract}

\vspace{-1em}
\section{Introduction}
\vspace{-1em}
The volume of research publications, particularly in AI-related fields, has accelerated dramatically due to the accessibility of modern academic workflows. This surge has made it increasingly difficult for genuinely novel work to stand out, as incremental papers often blend into the growing literature. Manual novelty assessment is time-consuming, subjective, and difficult to scale, motivating automated methods for estimating the originality of research ideas. Our goal is to develop a model that estimates and ranks the \textit{conceptual novelty} of AI research papers, providing a data-driven, consistent signal of originality. Such a system may help identify unconventional research directions and highlight submissions that introduce genuinely new ideas rather than minor variations.

We evaluate conceptual novelty using semantic information from a paper's title and abstract, along with its similarity to prior literature. To operationalize this, we explore two task formulations: (1) \textbf{binary classification}, which predicts absolute novelty from supervised examples, and (2) \textbf{pairwise comparison}, which learns relative novelty through pairwise preference signals. We fine-tune Qwen3-4B-Instruct-2507~\cite{qwen} and SciBERT~\cite{scibert} on both tasks, and benchmark against GPT-5.1~\cite{openai_gpt51} in a zero-shot setting to analyze the impact of model scale and supervision.

Our contributions are threefold:
(1) we formalize \textit{conceptual novelty estimation} as \textbf{context-aware conceptual deviation} from prior literature and instantiate it through binary classification and pairwise comparison tasks;
(2) we benchmark domain-specific, mid-sized fine-tuned, and zero-shot frontier models, finding that targeted fine-tuning of compact models outperforms zero-shot usage of significantly larger models;
(3) we deploy the best-performing model as an interactive system for real-time novelty scoring and retrieval, demonstrating practical usability for literature exploration.

Code is available on GitHub\footnote{\url{https://github.com/ZhengxuYan/NoveltyRank}}, and the system is deployed as a web application\footnote{\url{https://novelty-rank.vercel.app/}} for interactive exploration and community feedback.

\section{Related Work}

\textbf{Document Representation and Scholarly Retrieval.}
Transformer-based representations have become central to scientific document indexing and retrieval. SPECTER and SPECTER2 map papers into embedding spaces aligned with citation intent, enabling semantic search and clustering~\cite{specter2}. Systems such as \textit{arXiv Sanity Preserver}~\cite{karpathy2018arxiv}, Semantic Scholar, and topic-based recommenders surface relevant works via semantic similarity or metadata signals. However, these systems emphasize relevance rather than conceptual originality. Our framework is complementary: novelty scores can be layered on top of retrieval pipelines to surface unconventional or frontier ideas within a topic.

\textbf{Novelty and Originality Estimation.}
Prior work models novelty as deviation from historical literature via supervised classification~\cite{ghosal2018tapdlnd10corpus}, semantic redundancy detection~\cite{ghosal-etal-2018-novelty}, or network-based atypicality in citation/co-author graphs~\cite{AMPLAYO2018542}. Outlier-centric approaches estimate novelty by density or distance metrics in embedding space, e.g., fastText + LOF for biomedical titles~\cite{JEON2023101450}. These methods rely heavily on citation structures or handcrafted statistical assumptions.
In contrast, our approach incorporates transformer-based representations with retrieval-anchored contextual signals, enabling the model to assess \textbf{context-aware conceptual deviation} rather than mere semantic proximity. We further study how task formulation and model scale shape novelty prediction performance.

\section{Dataset}
\subsection{Data Source and Labeling}
Our dataset combines web-scraped arXiv entries with the public ICLR 2017--2025 dataset~\cite{gonzalez2024learning}, totaling \textbf{60,294 papers} published between 2023 and 2025. The corpus includes 50,442 randomly sampled arXiv papers and 9,852 papers accepted to top-tier venues across six domains (AI, ML, CV, Robotics, NLP, and Cryptography). For each entry, we retain metadata including paper ID, publication date, title, authors, and abstract.

Following prior work using venue acceptance as a heuristic signal for originality, we adopt \textbf{conference acceptance as a proxy label} for conceptual novelty: accepted papers are assigned label $1$ (positive) and randomly sampled arXiv papers are assigned label $0$ (negative). To prevent temporal leakage, we perform a chronological split: models are trained on papers from 2024 to early 2025 and evaluated on papers published after March 15, 2025. This setup reflects a real-world deployment scenario where novelty estimation is applied to future, unseen submissions.

\subsection{Document Representations}
We encode each paper using \textbf{SPECTER2}~\cite{specter2}, a transformer model trained for scientific document representation. Titles and abstracts are mapped to embedding vectors via two model heads with different semantic emphases:
\begin{itemize}
    \item \textbf{Classification Embedding} — captures a paper's semantic content for downstream prediction.
    \item \textbf{Proximity Embedding} — optimized with citation-based contrastive learning to reflect relational distance between papers in citation space.
\end{itemize}
These representations provide complementary signals: the former models internal semantics, while the latter situates the paper within the scientific landscape.

\subsection{Neighborhood-based Features (Retrieval-Augmented)}
To incorporate contextual signals beyond intrinsic document semantics, we adopt a retrieval-augmented design. Using proximity embeddings, each paper retrieves its top-$10$ most similar prior works via \textbf{Faiss}~\cite{faiss}, restricted to strictly earlier publication dates to avoid future information leakage. From these retrieved neighborhoods, we compute statistical features such as similarity aggregates (e.g., max/mean similarity) and deviation profiles that summarize how atypical a paper is relative to its closest historical neighbors. These retrieval-derived features are concatenated with the base document embeddings, enabling the model to assess novelty through both semantic representation and contextual deviation from prior literature.

\vspace{-0.5em}
\section{Models}
\vspace{-0.5em}
To implement the NoveltyRank framework, we formulate novelty estimation through two distinct tasks both aligned with the goal of evaluating scientific innovation: \textbf{Binary Classification} (evaluating the absolute novelty of a single paper) and \textbf{Pairwise Comparison} (assessing relative novelty between two papers). These tasks differ in input structure and supervision.

To benchmark performance across computational scales, we experiment with each task formulation on three models listed below following a specific logic: as model size decreases, the degree of task-specific adaptation increases. Such comparison determines whether smaller models with targeted parameter updates can match or surpass larger general-purpose models, verifying the feasibility of lightweight deployment.

\begin{itemize}
    \item \textbf{GPT-5.1 (Large-Scale / Zero-shot):} 
    As a frontier model, GPT-5.1 serves as the upper-bound baseline. It is accessed via API and evaluated in zero-shot without parameter updates.
    
    \item \textbf{Qwen3-4B (Mid-Scale / LoRA Tuning):} 
    Qwen3-4B balances size and adaptability. We apply a two-stage fine-tuning with Supervised Fine-Tuning (SFT) followed by Direct Preference Optimization (DPO)~\cite{dpo}, using LoRA \cite{lora} to update a small subset of parameters while freezing the backbone.
    
    \item \textbf{SciBERT (Small-Scale / Layer-Frozen Fine-Tuning):} 
    SciBERT represents the compact, domain-specific model. We adopt a multimodal approach by concatenating the standard SciBERT [CLS] token with pre-computed SPECTER2 embeddings and similarity features. To preserve scientific linguistic knowledge, the lower 8 encoder layers are frozen, and only upper layers and task-specific heads are fine-tuned.
\end{itemize}

Our models are trained in PyTorch~\cite{paszke2019pytorch} with the HuggingFace Transformers library~\cite{wolf2019huggingface}.

\vspace{-1em}
\section{Task Formulation 1: Binary Novelty Classification}
\vspace{-0.5em}
For binary novelty classification, we formulate the task as a supervised learning problem where the model primarily uses papers’ title, abstract, and similarity scores to predict binary novelty label (0 or 1). The objective is to learn general patterns and absolute criteria from established novel papers to assess the novelty of unseen samples.
\vspace{-0.5em}

\subsection{Qwen3-4B: SFT and DPO Fine-tuning}\label{subsec:Qwen_class}
\vspace{-1em}
\begin{figure}[ht]
    \centering
    \includegraphics[width=0.9\linewidth]{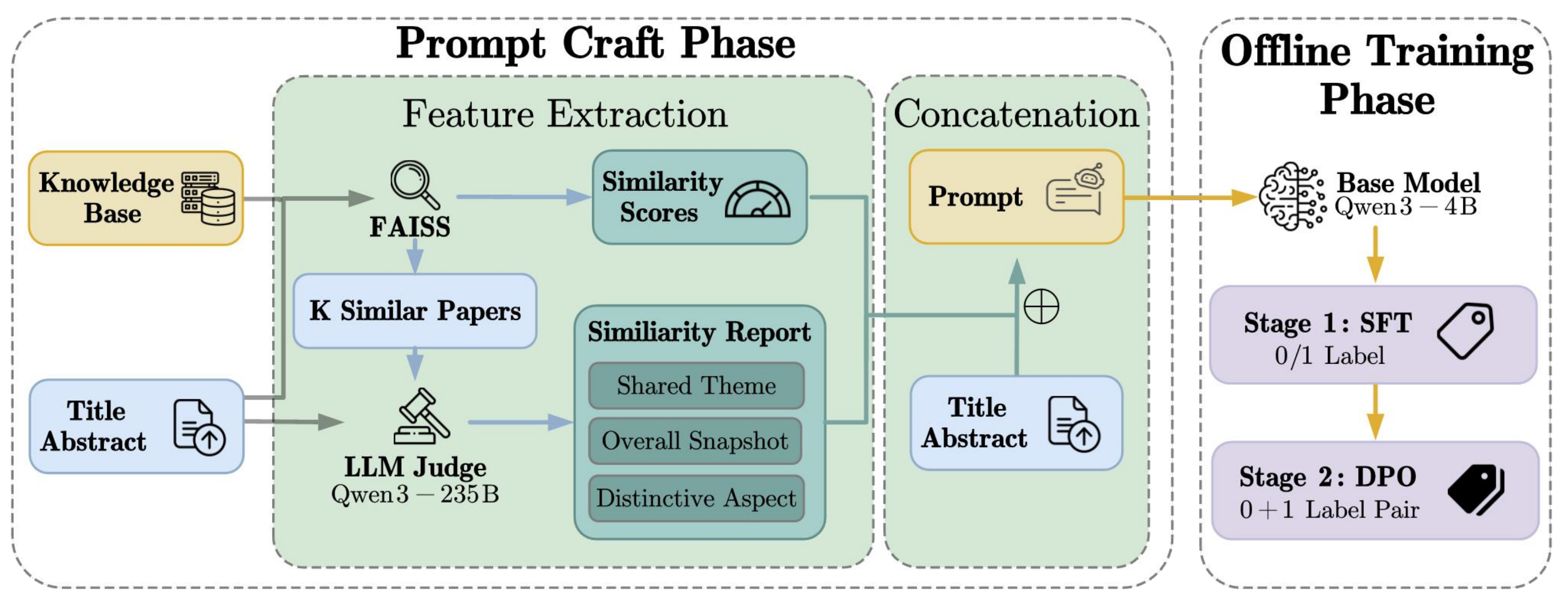}
    \vspace{-2mm}
    \caption{Qwen3-4B Pipeline for Binary Classification}
    \label{Qwen_4B_pipeline}
\end{figure}

\vspace{-2mm}
The full pipeline of Qwen3-4B model for binary novelty classification task is illustrated in Fig.~\ref{Qwen_4B_pipeline}. Because novelty is inherently comparative, we augment this input with a \textbf{Similarity Report} generated by an LLM Judge, which summarizes overlaps and distinctive contributions relative to the top-$K$ similar papers retrieved via FAISS \cite{faiss}. This comparison provides an additional signal for decision-making process. 

Although a base Qwen3-4B model could serve as the judge, we use Qwen-235B to exploit its superior reasoning capabilities for higher-quality analysis. To optimize training efficiency, all similarity reports are pre-generated. We further improve judge performance through prompt engineering, including Chain-of-Thought (CoT) \cite{chain} and few-shot examples \cite{shot}. A prompt example is provided in the Appendix.

\vspace{-0.5em}
\paragraph{SFT} 
In SFT, we apply cross-entropy loss between the model’s generated tokens and ground-truth labels. This setup naturally yields a binary supervision signal, training the model to generate discrete outputs (0 or 1). We avoid prompting for continuous confidence scores (e.g., 0.85), as LLMs generate text tokens rather than mathematically grounded probabilities. The decimal outputs would be linguistic hallucinations, rendering the model unreliable for quantitative confidence estimation. 
\vspace{-0.5em}
\paragraph{DPO} 
For DPO, we initialize the model using the SFT checkpoint to ensure training stability. We construct preference pairs $(y_{chosen}, y_{rejected})$ based on the ground truth of submission acceptance: if the chosen response is "1" and the rejected response is "0", the correct label is "1" (and vice versa). DPO then optimizes the model's likelihood to favor the correct classification over the incorrect one, further refining the model's ability to robustly distinguish novelty. 

Full hyperparameter configurations for SFT and DPO are provided in Table~\ref{tab:qwen-hp} in Appendix.

\vspace{-0.5em}
\subsection{Fine-tuned SciBERT}
We fine-tune the pretrained \texttt{scibert-scivocab-uncased} model for binary classification by integrating textual and metadata features. The input sequence combines the paper's title, abstract, and primary categories, separated by [SEP] tokens and truncated to a maximum length of 512. To capture semantic context beyond the input text, we concatenate the SciBERT [CLS] token output (768-dim) with three external feature vectors: the SPECTER2 classification embedding (768-dim), the proximity embedding (768-dim), and an aggregated embedding of the top-10 similar papers (768-dim). 

These features are fused into a multi-modal representation and passed through a custom classification head consisting of a three-layer feed-forward network ($2306 \to 512 \to 128 \to 2$) with ReLU activation and a dropout rate of 0.1. The model is trained using Cross-Entropy loss with an AdamW optimizer ($\eta = 2e^{-5}$) and a linear warmup scheduler for 5 epochs. Full hyperparameter configurations for the SciBERT encoder are provided in Table~\ref{tab:scibert_hyperparams} in Appendix.

\vspace{-0.5em}
\subsection{Evaluation Metrics for Binary Novelty Classification}
Given binary classification task and the class imbalance in our dataset, we evaluate model performance primarily using Precision, Recall, and F1-Score, which better reflect effectiveness on the minority (novel) class. Accuracy is also reported for completeness, but it may be misleading in cases where the model predicts predominantly negative labels. 

\subsection{Results and Discussion}
Table \ref{tab:classification_metrics} presents the performance of binary classification task on the test set, which consists of 10,889 examples with a highly imbalanced distribution (1,358 positives, approximately 12.5\%). The discussion follows.

\begin{table}[h!]
\centering
\vspace{-3mm}
\caption{Test Performance of Binary Classification (n=10,889)}
\begin{tabular}{ccccc}
\toprule
\textbf{Model} & \textbf{Accuracy} & \textbf{Precision} & \textbf{Recall} & \textbf{F1-score} \\
\midrule
\textbf{GPT-5.1} & 0.242 & 0.120 & 0.986 & 0.215 \\
\textbf{SFT Qwen3-4B} & 0.627 & 0.194 & 0.632 & 0.297 \\
\textbf{DPO Qwen3-4B} & 0.612 & 0.205 & 0.735 & 0.321 \\
\textbf{Fine-tuned SciBERT} & 0.744 & 0.187 & 0.313 & 0.234 \\
\bottomrule
\end{tabular}
\label{tab:classification_metrics}
\vspace{-3mm}
\end{table}

\paragraph{Performance of Large-Scale Models}
GPT-5.1 exhibits a strong "generosity bias" (Recall 0.986, Precision 0.120). Without specific training, the model tends to label nearly all papers as novel, failing to establish a rigorous boundary to filter out incremental works.
\vspace{-3mm}
\paragraph{Effectiveness of DPO}
DPO improves upon the SFT baseline (F1 0.321 vs. 0.297) primarily by boosting Recall (0.735 vs. 0.632). This confirms that preference optimization effectively encourages the model to identify valid novelty signals actively rather than defaulting to safe predictions.
\vspace{-3mm}
\paragraph{The Accuracy Paradox}
Despite higher accuracy (e.g., SciBERT's 74.4\%), fine-tuned models suffer from dataset imbalance. Low recall (especially SciBERT's 0.313) indicates the models collapse into conservative classifiers, minimizing loss by over-predicting the majority "non-novel" class.
\vspace{-3mm}
\paragraph{Insights on Absolute Novelty}
The suboptimal F1-scores across all models suggest that novelty is inherently relative, not absolute. Learning a crisp binary boundary from isolated inputs is difficult due to vague definitions. This limitation motivates our shift to a pairwise comparison formulation, where novelty is assessed relatively rather than absolutely.

\begin{figure}[t]
    \centering
    \includegraphics[width=0.9\linewidth]{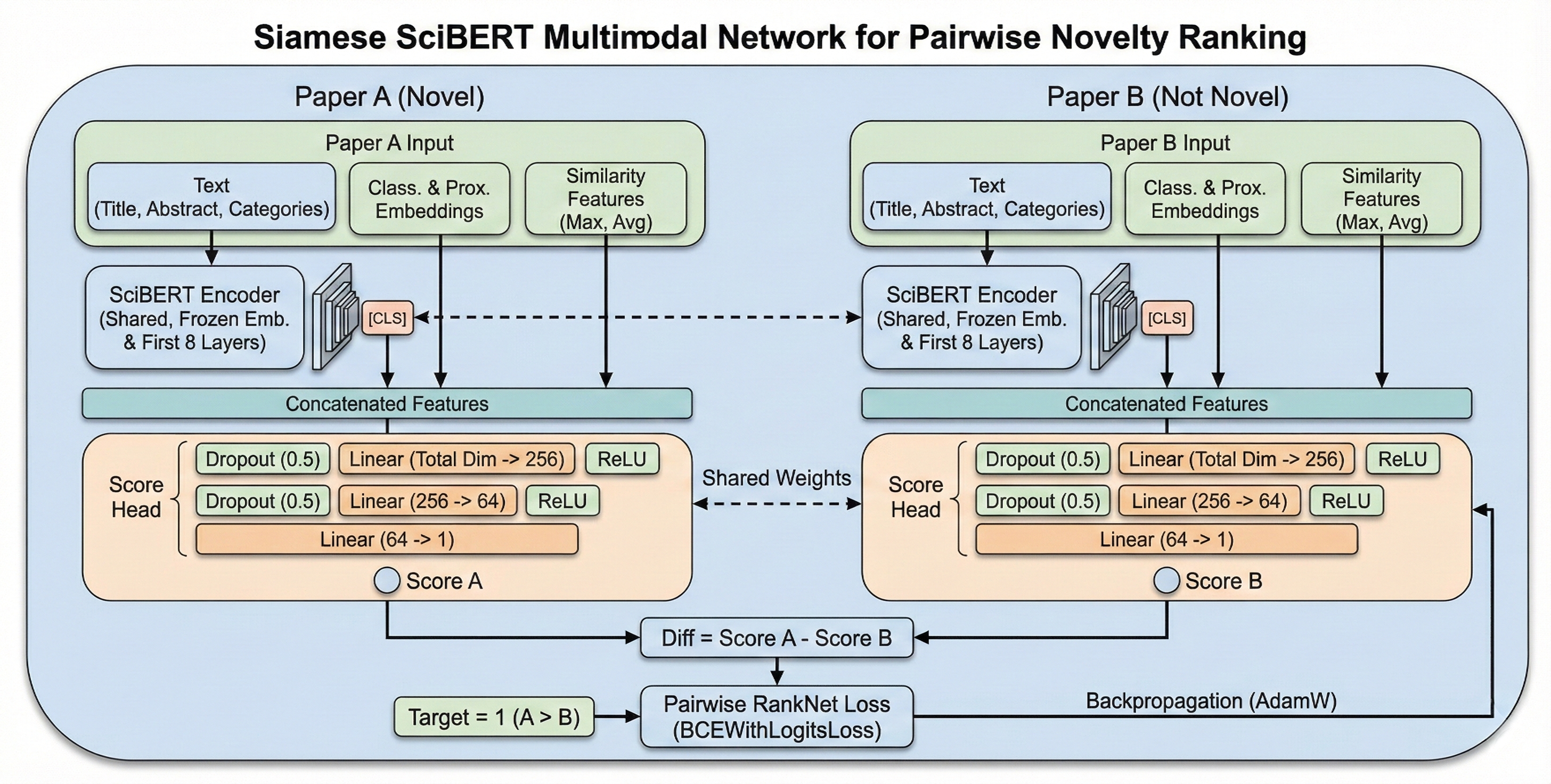}
    \caption{Siamese SciBERT Network}
    \label{scibert_framework}
\end{figure}

\vspace{-0.5em}
\section{Task Formulation 2: Pairwise Novelty Comparison}
\vspace{-0.5em}
For pairwise novelty comparison, the model jointly evaluates two papers to determine which one exhibits greater originality. Unlike binary classification, which learns absolute novelty criteria, this formulation focuses on relative patterns of innovation, enabling the model to discriminate novelty based on direct comparison.

Although each paper uses the same feature set as in the binary classification task, we construct comparison pairs to reflect the comparative nature of the objective. For each novel paper, we sample
a non-novel counterpart from the same domain to ensure meaningful contrast. To address class imbalance, we generate five such comparison pairs per positive paper by randomly sampling
multiple negative examples. To prevent positional bias (e.g., model systematically favoring the first option), we randomly shuffle the order of two papers within the pair during training.

\vspace{-0.5em}
\subsection{Qwen3-4B: SFT and DPO Fine-tuning}
\vspace{-0.5em}

The same prompt-engineering methods and the SFT and DPO methods described in Section~\ref{subsec:Qwen_class} apply here. However, the supervision signal shifts from binary labels (0 or 1) to positional indicators (Paper A or Paper B). This allows the model to adapt the same optimization pipeline to a relative comparison setting.

\vspace{-0.5em}
\subsection{Siamese SciBERT Network}To support the pairwise comparison task, we implement a Siamese network architecture with shared weights (as shown in Fig.~\ref{scibert_framework}). The model takes a pair of papers $(P_A, P_B)$ as input, processing each through identical SciBERT encoders to produce scalar novelty scores $s_A$ and $s_B$. To improve generalization and prevent overfitting on the smaller dataset of pairs, we freeze the embeddings and the first 8 transformer layers, fine-tuning only the top 4 layers of the encoder.

Similar to the classification task, the textual representation is concatenated with classification and proximity embeddings. This combined vector is fed into a scoring head ($2306 \to 256 \to 64 \to 1$) with a higher dropout rate of 0.5 to act as a regularizer. The network is optimized using RankNet loss, formulated as binary cross-entropy on the score difference $\sigma(s_A - s_B)$, effectively maximizing the likelihood that the novel paper is scored higher than its non-novel counterpart.

\vspace{-1em}
\subsection{Evaluation Metrics for Pairwise Novelty Comparison}
For the novelty comparison task, we evaluate performance using Pairwise Agreement, the proportion of pairs in which the model correctly identifies the more novel paper. Unlike the training phase which relied on random sampling (1:5 ratio), evaluation employs a dense pairing strategy: every positive paper is paired with all available negative samples in the same domain. This exhaustive matching eliminates sampling variance and maximizes the utilization of the test set for a robust assessment.

\subsection{Results and Discussion}

We constructed {9,531 testing pairs spanning six distinct domains. Table \ref{tab:overall_agreement} presents the \textbf{aggregate test agreement rates}, while Figure \ref{fig:category_analysis} provides a detailed breakdown of the test performance by domain and illustrates its distribution within the training set.

\begin{table}[h!]
\centering
\vspace{-3mm}
\caption{Performance of Pairwise Comparison (n=9,531)}
\begin{tabular}{lcccc}
\toprule
\textbf{Metric} & \textbf{GPT-5.1} & \textbf{SFT Qwen3-4B} & \textbf{DPO Qwen3-4B} & \textbf{FT SciBERT} \\
\midrule
\textbf{Agreement} & 0.583 & 0.739 & 0.741 & 0.753 \\
\bottomrule
\end{tabular}
\label{tab:overall_agreement}
\vspace{-3mm}
\end{table}

\paragraph{Efficacy of Fine-Tuning and Task Formulation} 
The results demonstrate that task-specific fine-tuning offers a clear advantage over generalized large-scale models. While the GPT-5.1 baseline achieved only marginal agreement (0.583), all fine-tuned models performed substantially better, led by SciBERT (0.753) and DPO-tuned Qwen3-4B (0.741).

These significant gains validate two key points: First, that domain-adapted models outperform frontier models for our task, despite their smaller size. Second, the high, consistent agreement rates confirm the effectiveness of the pairwise comparison formulation, which provides a clearer, more actionable training signal than binary classification formulation.

\begin{figure}[ht]
    \centering
    \includegraphics[width=0.8\linewidth]{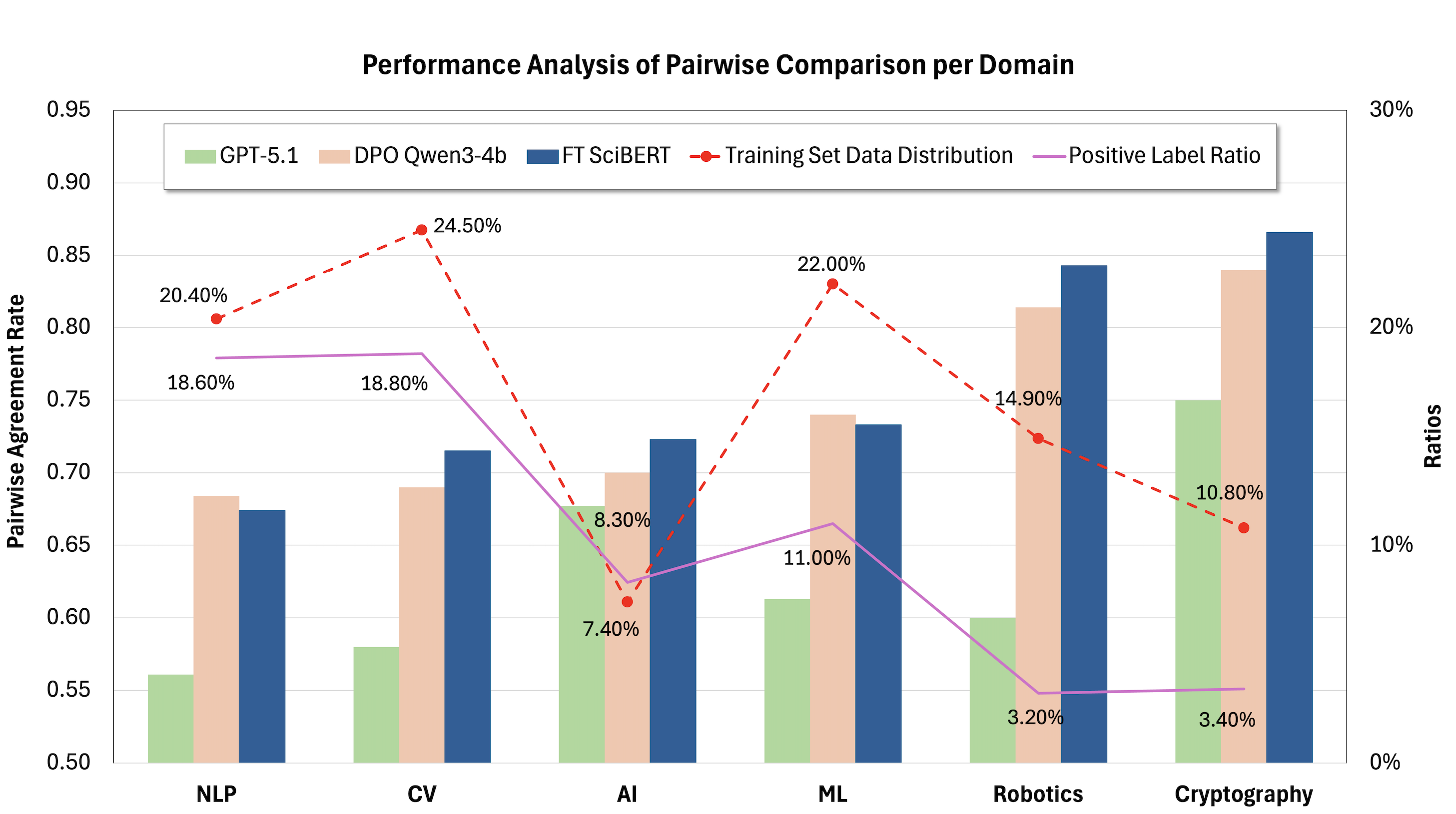}
    \caption{Comparison performance by domain. The bar chart shows each domain's test agreement rates alongside the proportion of positive labels and the category distribution within the training set.}
    \label{fig:category_analysis}
\end{figure}
\vspace{-4mm}



\paragraph{The "Consistency over Quantity" Paradox}
A closer examination of category-specific performance (Figure~\ref{fig:category_analysis}) reveals an inverse relationship between data volume and model accuracy. Models achieve their highest agreement in small, imbalanced fields such as Robotics (0.81--0.84) and Cryptography, even though these categories constitute only a minor portion of the training data. In contrast, large fields like Machine Learning yield lower agreement.

This suggests that pair consistency, rather than dataset size, is the primary driver of optimization quality. Niche domains tend to be semantically compact, producing pairs where “novel’’ and “non-novel’’ examples are closely aligned and easier to compare. Large heterogeneous domains, however, generate pairs spanning divergent subtopics, reducing semantic coherence and making relative novelty harder to judge. Overall, these results indicate that for pairwise comparison tasks, well-structured, semantically aligned pairs matter more than raw data scale.

\vspace{-0.5em}
\section{Conclusion}
\vspace{-0.5em}

This project introduces NoveltyRank, a system to evaluate the conceptual originality of AI papers. We find that pairwise comparison formulation offers a significantly cleaner and more effective learning signal than binary classification formulation. Moreover, we investigate three model scales and find that the cost-effective, domain-specific SciBERT, fine-tuned under the comparison setting, achieves best results. This demonstrates that comparative structure with small-model fine-tuning provides a more efficient and effective solution than increasing model size. 


\section*{Contributions}
All team members contributed to project deliverables.

\textbf{Jason Yan}: Conceived the initial project idea and developed the data scraping pipeline. Implemented the Siamese SciBERT multimodal network, and engineered the user interface for demonstration.

\textbf{Christine Li}: Generated SciBERT and Specter2 text embeddings and performed the FAISS similarity search. Conducted the downstream similarity computation and performed data validation tasks.

\textbf{Yuming Feng}: Architected the core comparison task formulation and implemented the entire end-to-end training pipeline for Qwen3-4B using both SFT and DPO methodologies. Primarily authored the Interpretation and Discussion of the experimental results.

\section*{Acknowledgments}
We would like to express our sincere gratitude to \textbf{Cody Ho} and \textbf{Ryan Lian} for their valuable feedback and guidance throughout this project. We are also grateful to \textbf{Andrew Ng} and \textbf{Kian Katanforoosh} for the insightful course content and discussions that shaped our understanding of applied machine learning. Finally, we thank \textbf{Thinking Machines Lab} for providing access to \textit{Tinker} compute resources that supported our experiments.

\appendix
\section*{Appendix}

\subsection*{A. Dataset}

The complete dataset created for this project is available on HuggingFace\footnote{\url{https://huggingface.co/datasets/JasonYan777/novelty-rank-with-similarities}} 
to support reproducibility and future research.

\subsection*{B. Hyperparameters}
Key hyperparameters for Qwen3-4B (SFT and DPO) and SciBERT include:
\begin{itemize}
    \item Number of epochs, batch size, and learning rate
    \item Optimization settings
    \item LoRA rank and scaling parameters (for Qwen3-4B)
\end{itemize}

\subsection*{B.1 Qwen-4B Hyperparameters}
\vspace{-1em}
\begin{table}[H]
\centering
\small
\caption{Hyperparameters for Qwen3-4B across classification and comparison tasks (SFT and DPO).}
\begin{tabular}{lcccc}
\toprule
\textbf{Hyperparameter} & \textbf{Class-SFT} & \textbf{Class-DPO} & \textbf{Comp-SFT} & \textbf{Comp-DPO} \\
\midrule
Learning rate & 2e-5 & 1e-6 & 3e-5 & 1.5e-6 \\
Batch size & 256 & 128 & 64 & 64 \\
Epochs & 4 & 1 & 10 & 4 \\
Max sequence length & 4096 & 1024 & 4096 & 1024 \\
LR scheduler & linear & linear & linear & linear \\
\midrule
Optimizer & AdamW & AdamW & AdamW & AdamW \\
Adam $\beta_1$ & 0.9 & 0.9 & 0.9 & 0.9 \\
Adam $\beta_2$ & 0.95 & 0.95 & 0.95 & 0.95 \\
Adam $\epsilon$ & 1e-8 & 1e-8 & 1e-8 & 1e-8 \\
\midrule
LoRA rank ($r$) & 32 & 32 & 32 & 32 \\
\midrule
DPO $\beta$ (only for DPO) & -- & 0.1 & -- & 0.1 \\
\bottomrule
\end{tabular}
\label{tab:qwen-hp}
\end{table}

\subsection*{B.2 SciBERT Hyperparameters}
\vspace{-1em}
\begin{table}[H]
\centering
\small
\caption{Hyperparameters for SciBERT across binary classification and pairwise comparison tasks.}
\begin{tabular}{lcc}
\toprule
\textbf{Hyperparameter} & \textbf{Binary Class.} & \textbf{Pairwise Comp.} \\
\midrule
Learning rate & 2e-5 & 1e-5 \\
Batch size & 32 & 64 \\
Epochs & 5 & 5 \\
Max sequence length & 512 & 512 \\
Weight decay & 0.01 & 0.1 \\
Warmup ratio & 0.1 & 0.1 \\
\midrule
Optimizer & AdamW & AdamW \\
Gradient accumulation & 1 & 2 \\
Dropout rate & 0.1 & 0.5 \\
Frozen layers & None & Embeddings + Layers 0-7 \\
\bottomrule
\end{tabular}
\label{tab:scibert_hyperparams}
\end{table}

\subsection*{C. Prompt Templates}

\subsubsection*{C.1 Binary Novelty Classification Prompt}

\begin{lstlisting}[basicstyle=\ttfamily\footnotesize, numbers=none]
########################
# System Prompt
########################
You are an expert AI researcher and senior conference reviewer (NeurIPS/ICLR level).
Your goal is to judge whether the submission introduces a conceptually novel idea.
Conceptual novelty captures fundamental shifts in scientific thinking.

---
### Conceptual Novelty Primer
Consider the following signals:
- Problem Formulation: Does it redefine an existing task or introduce a new one?
- Methodological Innovation: Does it propose a new class of algorithms or training paradigm?
- Theoretical Insight: Does it deliver a unifying or surprising theoretical lens?
- Cross-Disciplinary Import: Does it import a transformative idea from another domain?
Incremental tweaks (hyperparameters, surface-level architecture edits, dataset swaps) are not novel.

---
### Reference Decisions
Example 1:
Title: Differentiable Logic for Robotics
Abstract: Introduces a framework that composes continuous control policies with symbolic logic programs to enable reasoning-guided motion planning.
Similarity scores: max=0.61 | avg=0.48
Reasoning: Combines two previously disjoint paradigms (continuous control and symbolic reasoning) into a unified differentiable architecture (Novel).
Output: 1
Example 2:
Title: Better Hyperparameters for BERT Fine-Tuning
Abstract: Reports extensive sweeps over learning rates and batch sizes for BERT on GLUE benchmarks.
Similarity scores: max=0.89 | avg=0.81
Reasoning: Purely empirical tuning without a new formulation or architecture (Not Novel).
Output: 0
Example 3:
Title: Physical Priors for Diffusion Models
Abstract: Incorporates symbolic conservation laws into diffusion model training to improve controllable generation.
Similarity scores: max=0.67 | avg=0.58
Reasoning: Introduces a cross-disciplinary inductive bias that reshapes the generative objective (Novel).
Output: 1

########################
# User Prompt
########################
---
### Paper Metadata
Title: {title}
Primary Category: {category}
Abstract: {abstract}
Max similarity to prior work: {max_sim}
Average similarity to prior work: {avg_sim}
---
### Similarity Report (Aggregated)
{similarity_report}
---
### Decision Instructions
1. Synthesize the available evidence (abstract + similarity signals).
2. Decide whether the work represents a conceptually novel contribution.
3. Output "1" if the paper is conceptually novel and likely to influence future research.
4. Output "0" if the contribution is incremental, derivative, or lacks conceptual novelty.
Respond with a single digit (0 or 1).
\end{lstlisting}

\subsubsection*{C.2 Pairwise Novelty Judgment Prompt}

\begin{lstlisting}[basicstyle=\ttfamily\footnotesize, numbers=none]
########################
# System Prompt
########################
You are an expert computer-vision researcher and senior conference reviewer (CVPR/ICCV/NeurIPS level).
Your goal is to compare the *conceptual novelty* of two computer-vision research papers (not just surface/benchmark improvements).

---
Conceptual Novelty Primer
Consider the following signals:
- Problem Formulation: Does it redefine an existing task or introduce a new one?
- Methodological Innovation: Does it propose a new class of algorithms or training paradigm?
- Theoretical Insight: Does it deliver a unifying or surprising theoretical lens?
- Cross-Disciplinary Import: Does it import a transformative idea from another domain?
Incremental tweaks (hyperparameters, surface-level architecture edits, dataset swaps) are not novel.

---
Step-by-step reasoning (use these as your guide and mention the strongest signal):
1) Extract the core technical idea from each paper's title and abstract.
2) Check whether the idea represents a new task, representation, learning paradigm, or major architectural shift.
3) Use similarity metrics as supportive evidence (high similarity tilts toward incremental), but prioritize conceptual signals (new objective, representation, or theory).
4) Choose which paper is more conceptually novel; answer only with 'A' or 'B'.

--- EXAMPLES
Example 1:
Paper A: Introduces Vision Transformer (ViT), treats images as a sequence of patches and applies a pure transformer backbone, changing core architecture for vision.
Paper B: Reports small regularization and augmentation tweaks to ResNet training that marginally improve accuracy.
Reasoning: A introduces a new architectural paradigm for visual representation (Novel).
Output: A
Example 2:
Paper A: Proposes Neural Radiance Fields (NeRF), an implicit continuous 3D scene representation enabling view synthesis.
Paper B: Improves an existing multi-view stereo pipeline with a better post-processing filter.
Reasoning: NeRF introduces a fundamentally new representation and rendering paradigm (Novel).
Output: A
Example 3:
Paper A: Applies an off-the-shelf transformer to a small medical imaging dataset with minor changes.
Paper B: Proposes a new contrastive objective that aligns multi-resolution feature maps and demonstrates broad transfer across many vision tasks.
Reasoning: B defines a new learning objective with broad implications -> Novel.
Output: B

########################
# User Prompt
########################
---
### Paper A
Title: {titleA}
Primary Category: {categoryA}
Abstract: {abstractA}
Max similarity to prior work: {max_simA:.4f}
Average similarity to prior work: {avg_simA:.4f}
---
### Paper B
Title: {titleB}
Primary Category: {categoryB}
Abstract: {abstractB}
Max similarity to prior work: {max_simB:.4f}
Average similarity to prior work: {avg_simB:.4f}
---
Output only the single letter 'A' or 'B'.
"""
\end{lstlisting}

\medskip
\small
\bibliographystyle{unsrt}
\bibliography{refs}

\end{document}